\documentclass[runningheads]{llncs}
\usepackage{graphicx}
\usepackage{wrapfig}
\usepackage{subcaption}
\usepackage{float}
\usepackage{amsmath}
\usepackage{mathtools}
\usepackage[english]{babel}
\usepackage{cite}
\usepackage{textcmds}
\usepackage[utf8]{inputenc}
\usepackage{lineno}

\begin{document}
\title{Discrete models of continuous behavior of collective adaptive systems}
\titlerunning{Discrete models of continuous behavior}
%
%

\author{Peter Fettke\inst{1,2}\orcidID{0000-0002-0624-4431} \and
Wolfgang Reisig\inst{3}\orcidID{0000-0002-7026-2810}}
\authorrunning{P. Fettke, W. Reisig}
%

\institute{German Research Center for Artificial Intelligence (DFKI), Saarbr\"ucken, Germany \\
\email{peter.fettke@dfki.de}\\ \and
Saarland University, Saarbr\"ucken, Germany \\ \and
Humboldt-Universität zu Berlin, Berlin, Germany \\  
\email{reisig@informatik.hu-berlin.de}}

\maketitle              
\begin{abstract}
Artificial ants are “small” units, moving autonomously on a shared, dynamically changing “space”, directly or indirectly exchanging some kind of information. Artificial ants are frequently conceived as a paradigm for collective adaptive systems. In this paper, we discuss means to represent continuous moves of “ants” in discrete models. More generally, we challenge the role of the notion of “time” in artificial ant systems and models. We suggest a modeling framework that structures behavior along causal dependencies, and not along temporal relations. We present all arguments by help of a simple example. As a modeling framework we employ \textsc{Heraklit}; an emerging framework that already has proven its worth in many contexts.

\end{abstract}

\section*{Introduction}

Some branches of informatics take processes in nature as a model for unconventional classes of algorithms. In particular, numerous variants of “swarm intelligence” have been and are being studied to a large scale, with specialized conference series, journals, etc., e.g. \textit{International Conference on Swarm Intelligence (ICSI)}, and \textit{International Journal of Swarm Intelligence Research (IJSIR)}. What many of these approaches have in common, is the assumption of a number of artificial ants, i.e. “small” units, moving autonomously around on a shared, dynamically changing “space”, directly or indirectly exchanging some kind of information. 

This kind of behavior needs an adequate representation, i.e. it must be modeled in a formal framework, as a basis for implementation, for proving correctness, for studies of complexity, and many other tasks.

In the following, we discuss fundamental assumptions and questions of modeling such systems. In particular, we discuss means to represent continuous moves of “ants” in discrete models. More generally, we challenge the role of the notion of “time” in artificial ant systems and models. We show that time-based models do not adequately represent the causal dependencies in such systems. Instead, we suggest a modeling framework that structures behavior along causal dependencies, and not along temporal relations. 

We present all arguments by help of a simple example, i.e., ants moving up and down a bar. This example already exhibits numerous fundamental problems of the area. As a modeling framework we employ \textsc{Heraklit} \cite{fettke2021modelling,fettke2021handbook}; an emerging modeling framework that already has proven its worth in many contexts, and shows its utility also in the context of artificial ants.

\section{Running example: ants on a bar}

Here we start out with an informal description of ants on a bar. We identify three kinds of events.

\subsection{The behavior of ants on a bar}
\begin{figure}[!tb]
\centering
{\includegraphics[trim={0cm 0cm 0cm 0cm},clip,width=\textwidth]{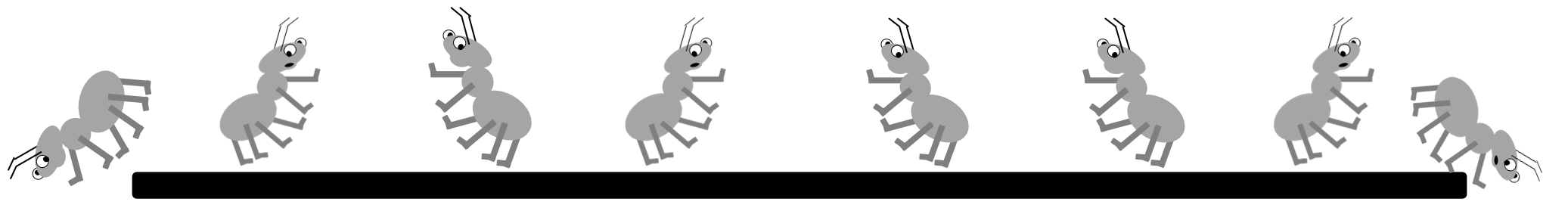}}
\caption{ants on a bar}
\label{fig:Ameisen}
\end{figure}

Assume a bar and some ants, moving up and down the bar, as in Fig.~\ref{fig:Ameisen}. An ant, moving towards the right end of the bar, and its right neighbor moving left, will eventually meet at some point on the bar. In this case, both ants swap direction of movement. A right moving ant without right neighbor will eventually drop down from the bar, and so will drop down a left moving ant without a left neighbor. Ants cannot get ahead of each other.

An initial state is a set of ants on the bar, where each ant is directed left or right. For every given initial state, the ants’ behavior is clear and unambiguous, generating a \textit{run}. The ants system is the set of all potential initial states and their runs. As the above description of the ants system is intuitively clear, it should not be too difficult to model it formally. However, there is much to say about this seemingly simple endeavor.  

\subsection{Events of the ants’ system\label{sec:1.2}}
We start with the quest of modeling single runs. Starting in a given initial state, three types of events may occur:
\begin{enumerate}
\item Two ants $i$ and $j$ meet. As a shorthand we write “a i j” for this event. \item An ant $i$ drops down on the left edge of the bar. As a shorthand we write “b\, i” for this event. 
\item An ant $i$ drops down on the right edge of the bar. As a shorthand we write “c\, i” for this event. 
\end{enumerate}
Occurrence of an event causes a fresh state.

\section{Conventional models of the ants’ behavior}

There are several more or less different ways to represent the behavior of ants, embedding events into a temporal framework, and structuring the behavior along the flow of time. We consider four such representations and show what they have in common.

\subsection{The continuous model of a run of the ants system\label{sec:2.1}}
\begin{figure}[!tb]
\centering
{\includegraphics[trim={0cm 0cm 0cm 0cm},clip,scale=.25]{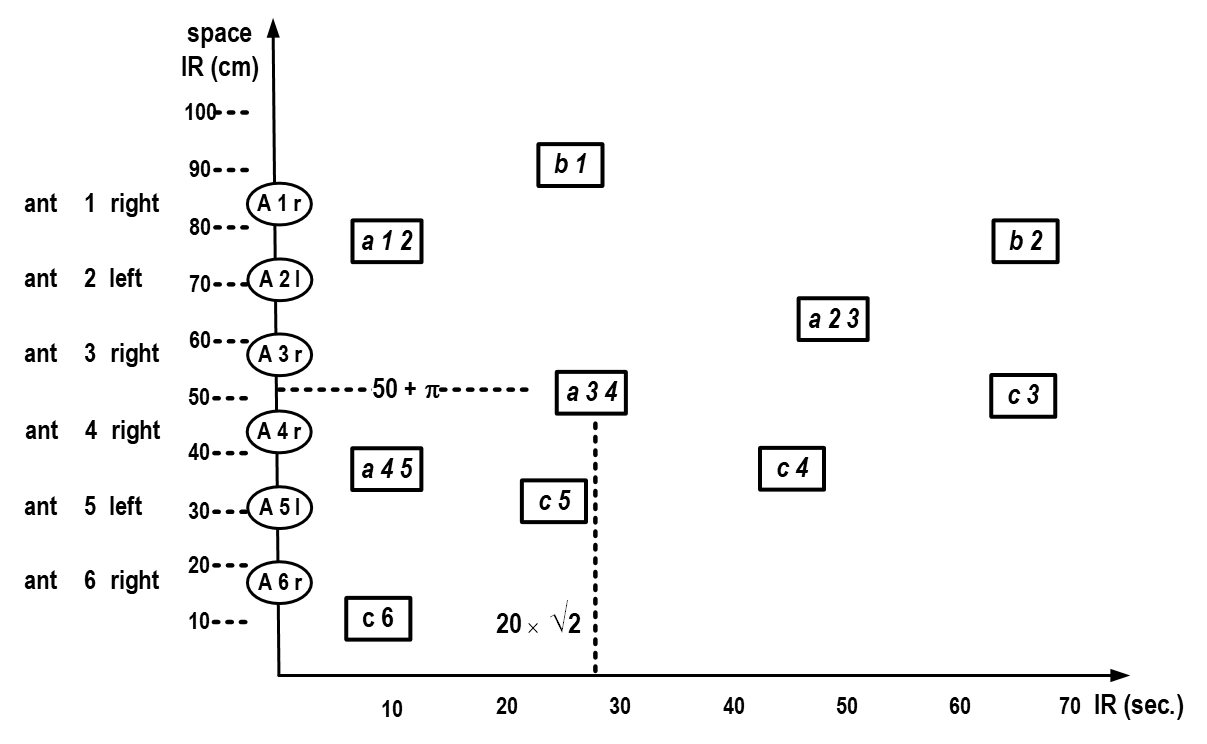}}
\caption{the continuous model of behavior}
\label{fig:klassisch}
\end{figure}

Fig.~\ref{fig:klassisch} models a behavior of the ants in the style of classical mechanical engineering: a Cartesian plain is spanned, with real numbers for time and for space in its $x$- and $y$-axis, respectively. An initial state $S_0$, reached at time $t = 0$, is outlined  as follows: For each ant $i$, an ellipse, inscribed $A\, i\, l$ or $A\, i\, r$, describes the initial position on the bar of ant $i$, as well as its orientation, left or right. For example, ant $1$ is the leftmost ant, oriented rightwards.

Fig.~\ref{fig:klassisch} shows a typical behavior of the events of the ants system. Any real number of the $x$-axis of Fig.~\ref{fig:klassisch} denotes a point in time, and any real number of the $y$-axis denotes a point in space on the bar, where an event may occur. For example, the “$r\, 6$” inscribed box indicates the event of ant number $6$ dropping down the right edge of the bar. The “$a\, 3\, 4$” inscribed box indicates that ants $3$ and $4$ meet. 

The model in Fig.~\ref{fig:klassisch} encompasses infinitely many, in fact more than countably many potential states apt to cope with “real time”. However, for the understanding of the functionality of a system, the aspect of “real time” is often irrelevant and makes little sense. An example would be the claim that ants number $3$ and number $4$ meet at location $50 + \pi$ ($= 50 + 3.14\dots$) cm on the bar, and $20 \times \sqrt{2}$ $(= 20 \times 1.47\dots)$ seconds after the initial state. Neither can anybody empirically measure such a claim, nor is this claim relevant for a proper understanding of a single run of the ants system, or the ants system as a whole. 

Starting at state $S_0$ as in Fig.~\ref{fig:klassisch}, the events as described in Sec.~\ref{sec:1.2} may occur at any point of time and any location on the bar. Hence, $S_0$ yields infinitely many different runs, where events occur at different times at different places on the bar. Nevertheless, each run exhibits some kind of structure: some events are definitely ordered. For instance, ant $4$ turns first right to meet ant $5$, and then left to meet ant $3$. Hence, event $a\, 4\, 5$ is a prerequisite for $a\, 3\, 4$. This observation gives rise to the \textit{causality requirement}:
\begin{equation}
\text{If event $e$ is a prerequisite for event $f$, then $e$ occurs before $f$.}
\end{equation}
It is the concept of causality, that structures the behavior, and captures its decisive properties.

Avoiding unnecessarily detailed aspects, informatics does mostly deals with \textit{discrete} models. A discrete model of behavior describes behavior by help of finitely many or countably many states and events, where an event updates a given state. This is aspired in different ways, most prominently the following three ones.

\subsection{The grid model\label{sec:2.2}}
\begin{figure}[!tb]
\centering
{\includegraphics[trim={0cm 0cm 0cm 0cm},clip,scale=.25]{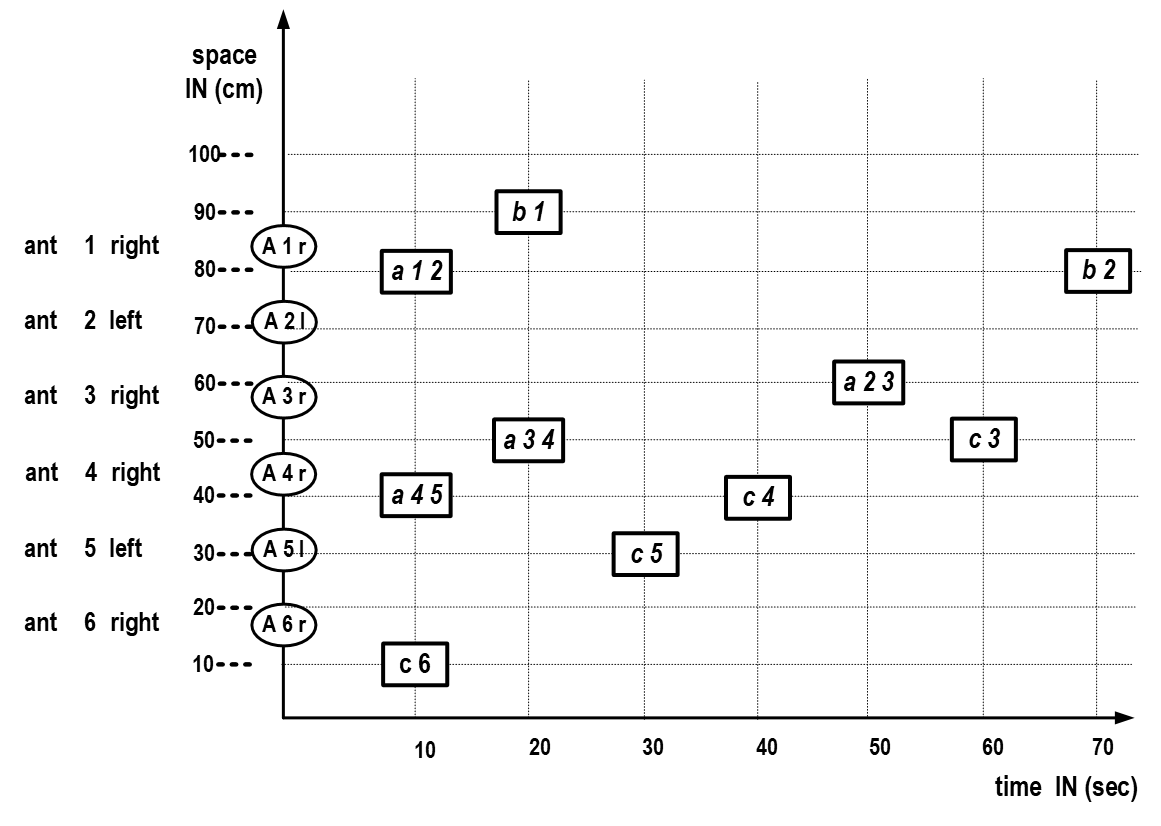}}
\caption{the grid model of behavior}
\label{fig:grid}
\end{figure}

The grid model spans a grid in the plane, e.g. the integer grid as in Fig.~\ref{fig:grid}. This grid cuts time and space into intervals. In each state of the system, each ant occupies a square of the grid. Upon an event, each ant is assumed to move right to the next time interval, and coincidently moving up or down to a neighbored  space interval, or to remain at its position. Consequently, ants meet or drop down from the bar at the grid’s crossings. This is what cellular automata and most implementations of ant algorithms do. Resnick suggests a programming language, based on the grid model \cite{resnik1994turtles}.

For a fixed initial state, the grid model does not define a unique run: depending on the choice of the grid, two events may be mapped to the same or to different points of the grid, i.e. two pairs of ants are considered as meeting in time coincidently (at different points of space on the bar), or in a sequence. Any such mapping of events onto grid points is acceptable, provided the causality requirement (1) holds in a slightly revised form:
\begin{gather}
\text{If $e$ is a prerequisite for $f$, then the time component of $e$}\nonumber \\
\text{must be smaller than the time component of $f$.}
\end{gather}

This is in fact the case in Fig.~\ref{fig:grid}. As a further example, the event $r\, 6$ indicates that ant number $6$ just drops down at the right edge of the bar, without meeting any other ant. Consequently, $r\, 6$ may be mapped to any of the seven grid points of time.

\subsection{The numbering model}
As a second possibility, one just numbers the ants’ meetings, such that the causality requirement holds in the varied formulation: 
\begin{gather}
\text{If $e$ is a prerequisite for $f$,}\nonumber \\
\text{then the number of $e$ must be smaller than the number of $f$.}
\end{gather}

\begin{figure}[!tb]
\centering
{\includegraphics[trim={0cm 0cm 0cm 0cm},clip,scale=.25]{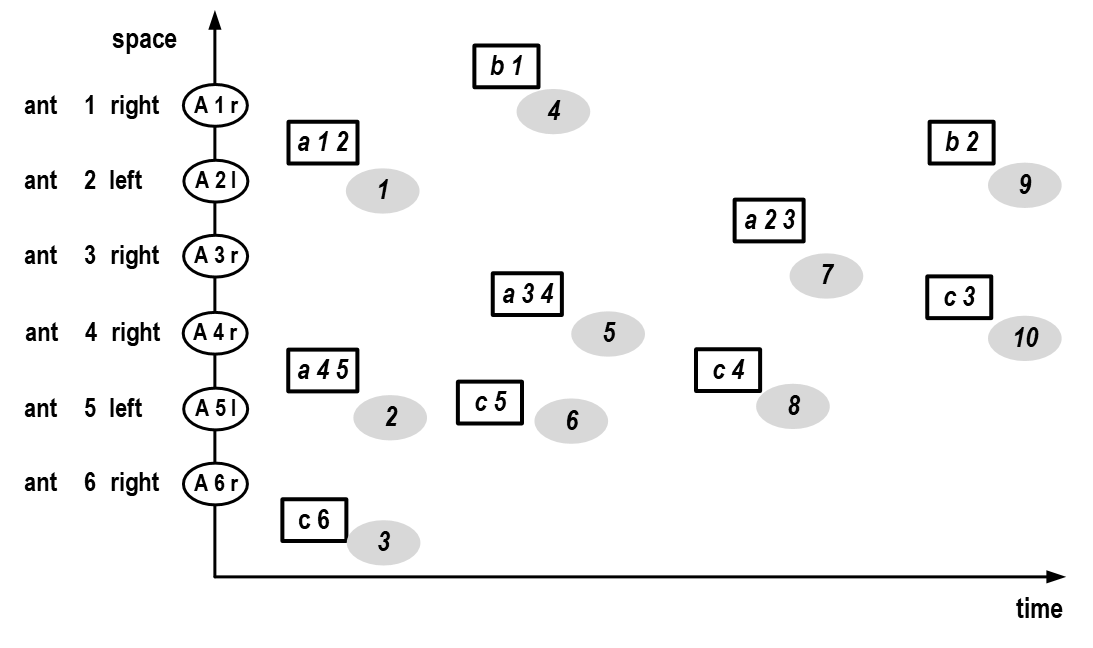}}
\caption{the numbering model}
\label{fig:numbering}
\end{figure}

For the above initial state $S_0$, Fig.~\ref{fig:numbering} shows an example for this model of behavior. There are various different acceptable numberings. For example, the numberings $2$ and $4$ of the events $a\, 4\, 5$ and $l1$ may be swapped.

\subsection{The lockstep model\label{sec:2.4}}
Behavior can also be modeled as to proceed in lockstep, i.e. in a sequence of steps, where each step is a set of events. Fig.~\ref{fig:lockstep} shows the ants’ behavior as a sequence of four steps, for the above initial state $S_0$. Again, the causality requirement must hold is a varied form:
\begin{gather}
\text{If $e$ is a prerequisite for $f$, then the step including $e$} \nonumber \\
\text{must occur in the lockstep sequence before the step including $f$.}
\end{gather}
Again, the running example generates not a unique lockstep sequence.

\subsection{The uniqueness problem}
For the given initial state $S_0$, the continuous model of Sec.~\ref{sec:2.1} as well as each of the above three discrete models cause various different runs. The runs all differ w.r.t. the time at which events occur. The assumption of time is the structuring principle of these runs. However, “time” come from outside the ant system. In forthcoming section 3, we strive at a discrete, yet unique model of the ants’ run, without outside structuring principles. But first we shed light onto the causality requirement (1) that occurs in some way in all models ((2), (3), (4)). This requirement induces an order on the events of a run.

\subsection{Weak orderings of events\label{sec:2.6}}
All above models of single runs (i.e. the continuous, the grid- the numbering- and the lockstep-model) order the evens of a run along the $x$-axis, representing the intuitive notion of time in various forms: as real numbers, discrete number intervals, integers, or just discrete, finite order. In all these models, events are either ordered in time, or they occur at the same time. This ordering is a \textit{weak ordering} on the events: for three events $e$, $f$, and $g$ always holds:
\begin{gather}
\text{If $e$ and $f$ are not ordered, and $f$ and $g$ are not ordered,} \nonumber\\
\text{then also $e$ and $g$ are not ordered.}
\end{gather}

For example, $a\, 1\, 2$, $a\, 4\, 5$ and $r\, 6$ occur at time $10$ in the grid, as well as in the first step of the lockstep run. Generally, a partial order, namely a transitive and irreflexive relation, is a weak ordering, if the complement of its symmetric closure is transitive. Consequently, any version of order that is motivated by the intuitive notion of “time” is a weak order, because in a temporally motivated order, unorder means “occurring at the same time”. And “occurring at the same time” is intuitively definitely transitive.

\section{The causal model of the ants’ runs}
Here we suggest a behavioral model that contrasts the above time-based model: A run is no longer structured along time, but along causality. It turns out, that the induced order is no weak order anymore.

\subsection{The order of events}
As an alternative to ordering events by temporal aspects, we start out with a closer look at the causality requirement, as stated in (1): “If event $e$ is a prerequisite for event $f$, then $e$ occurs before $f$”. Here, “to be a prerequisite of …” is a certainly a transitive relation: 
\begin{gather}
\text{If $e$ is a prerequisite for $f$, and if $f$ is a prerequisite for $g$,} \nonumber\\
\text{then $e$ is a prerequisite for $g$.}
\end{gather}

This implies that “to be a prerequisite of …” is a partial order. In the sequel, we denote it as an \textit{event order}.  Fig.~\ref{fig:event_order} shows the event order for the run of the ants system, starting in state $S_0$. Graphically, each arrow begins at a direct prerequisite of the event at the arrow’s end. Order on events induced by transitivity is not depicted. In Fig.~\ref{fig:event_order}, $c\, 5$ is unordered with $a\, 3\, 4$, as well as with $a\, 2\, 3$. But $a\, 3\, 4$ is ordered with (i.e. smaller than) $a\, 2\, 3$. Formulated more generally, if $e$ no prerequisite for $f$, and if $f$ is no prerequisite for $g$, then $e$ may very well be a prerequisite for $g$. In technical terms, this means: 
\begin{equation}
\text{In general, causal order is no weak ordering.}
\end{equation}

\begin{figure}[!tb]
\centering
{\includegraphics[trim={0cm 0cm 0cm 0cm},clip,scale=.25]{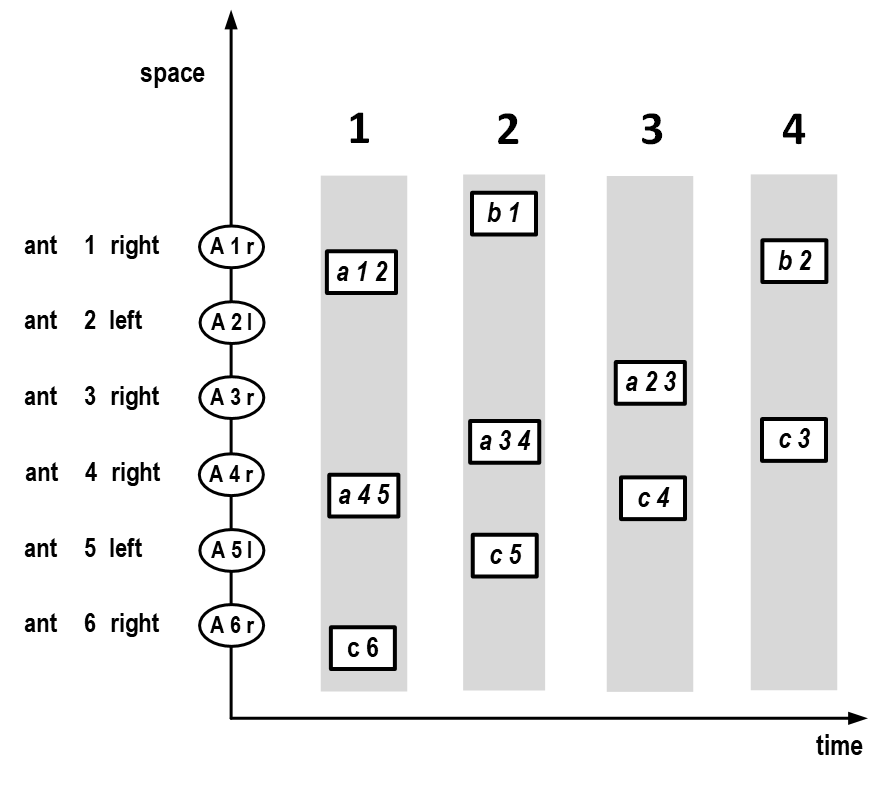}}
\caption{the lockstep model}
\label{fig:lockstep}
\end{figure}

This contrasts the temporally motivated orders of (1) to (4). 

We intend to employ event orders as a model of runs. This rises the problem of identifying states in a run. There is no way to insert global states into a partial order that is not weakly ordered. Nevertheless, wishing to embed single runs into behavioral system models, some aspects of states are inevitable.

\subsection{Local states and steps}
To include aspects of states into an event order, we pursue the idea of local states. For example, the initial state $S_0$ of the ants system, as in Figs.~\ref{fig:klassisch}, \ref{fig:grid}, \ref{fig:numbering}, and \ref{fig:lockstep} consists of local states, one for each ant.  

We construct local states as suggested in the framework of the \textsc{Heraklit} \cite{fettke2021handbook}: a local state is a proposition, usually a predicate $p$ together with an item or a tuple $t$, where $p$ applies to $t$. For example, let $A$ be the predicate “directed ants on the bar”. This predicate applies to the tuple $(1, r)$ in the initial state $S_0$ of the ants system. This kind of propositions is usually written $A(1,r)$. In graphical representations we skip the brackets. Hence, the initial state $S_0$ of the ants system is a set of local states, each of which is shaped “A\, i\, j”, with $i = 1, \dots , 6$, and $j \in \{l,r\}$. 

\begin{figure}[!tb]
\centering
{\includegraphics[trim={0cm 0cm 0cm 0cm},clip,scale=.25]{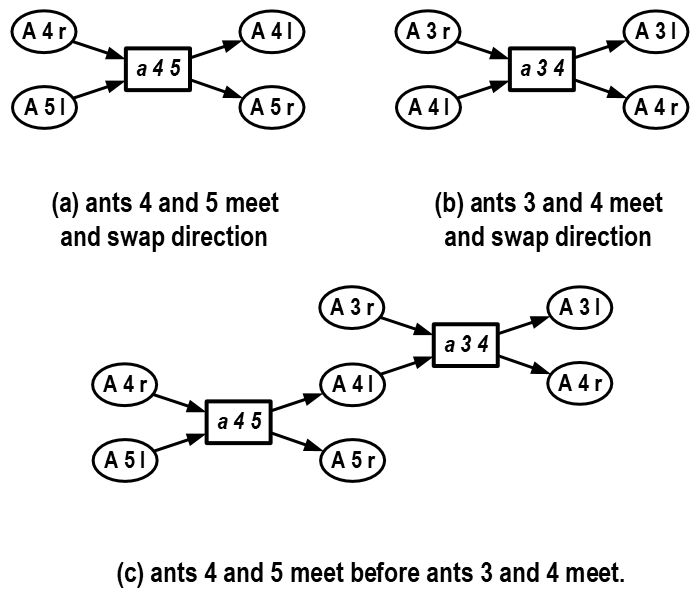}}
\caption{events and their composition}
\label{fig:single_events}
\end{figure}

A local step is an occurrence of an event, together with the event’s effect on local states. Each step updates some of the local states. For example, Fig.~\ref{fig:single_events} (a) shows the local step of meeting of ants $1$ and $2$, and their swapping of direction. This figure shows the cause and effect of event $a\, 4\, 5$ to the local states $A\, 4\, r$ and $A\, 5\, l$: they both are updated to $A\, 4\, l$ and $A\, 5\, r$, respectively. Any kind of global state is not necessary to specify event $a\, 4\, 5$. Correspondingly, Fig.~\ref{fig:single_events} (b) shows the meeting of ants $3$ and $4$. 

\begin{figure}[!tb]
\centering
{\includegraphics[trim={0cm 0cm 0cm 0cm},clip,scale=.25]{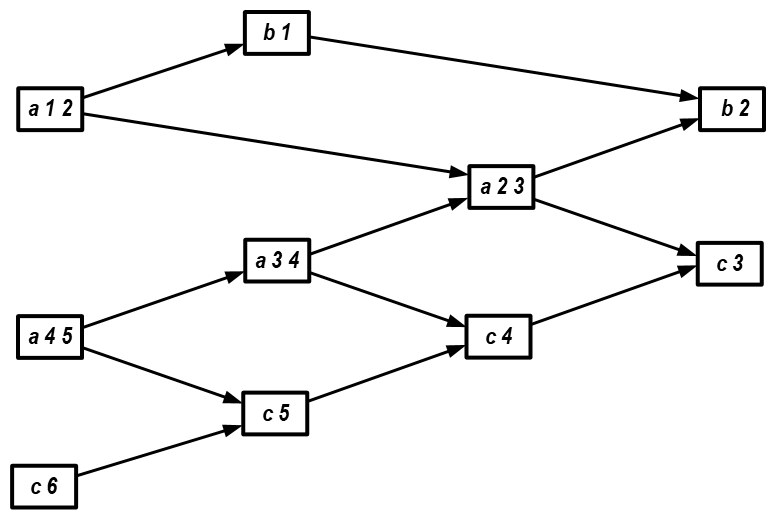}}
\caption{event order}
\label{fig:event_order}
\end{figure}

According to Fig.~\ref{fig:event_order}, the event $a\, 4\, 5$ is a prerequisite for $a\, 3\, 4$, because ant $4$
\begin{enumerate}
\item starts to the right (local state $A\, 4\, r$), 
\item then swaps to $A\, 4\, l$ (jointly with ant $5$), 
\item then returns back to $A\, 4\, r$, (jointly with ant $3$). 
\end{enumerate}
Fig.~\ref{fig:single_events} (c) shows the combined behavior of $a\, 4\, 5$ occurring before $a\, 3\, 4$.

\subsection{The run starting at state $S_0$}
\begin{figure}[!tb]
\centering
{\includegraphics[trim={0cm 0cm 0cm 0cm},clip,scale=.25]{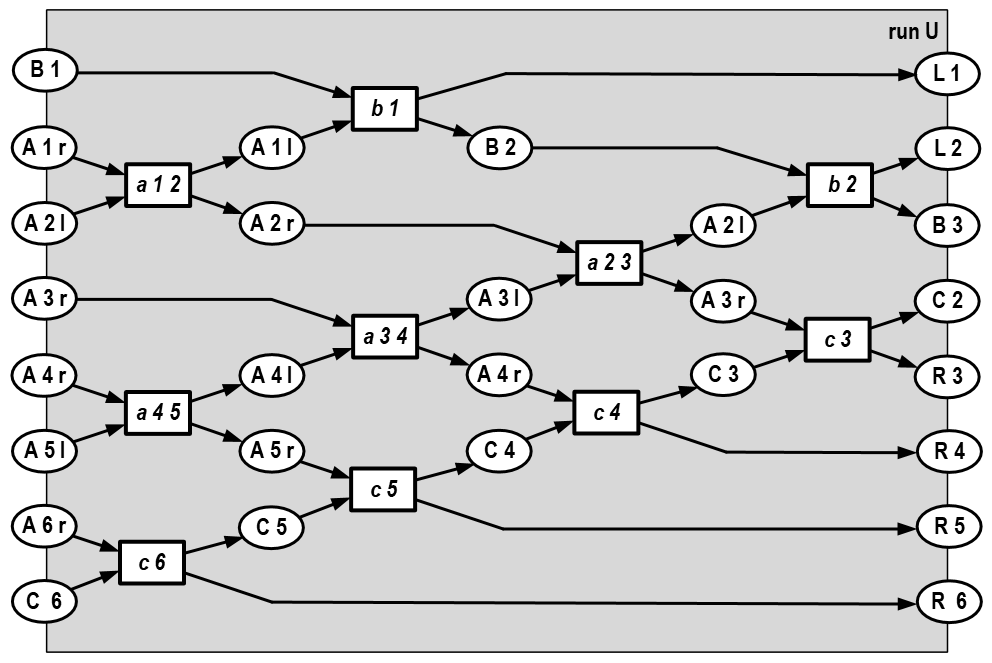}}
\caption{run $U$ of the ants system}
\label{fig:Gesamt-Ablauf}
\end{figure}

Fig.~\ref{fig:Gesamt-Ablauf} shows the run U of the ants system, starting in the initial state $S_0$. Besides the predicate “$A$”, the run $U$ employs further predicates, used to describe causes and effects of ants to drop down left and right: Local states $B\, i$ state that ant $i$ is the leftmost ant, hence $i$ is the next ant expected to drop down to the left; $L\, i$ states that ant $i$ has dropped down left. Accordingly, $C\, i$ stated that ant ant $i$ is the rightmost ant, hence next ant expected to drop down to the right; $R\, i$ states that ant $i$ has dropped down right. The gray background displays a \textsc{Heraklit} module, with interfaces on its left and right margin.

\subsection{Composing the run $U$ from the ants’ behavior}
\begin{figure}[!tb]
\centering
{\includegraphics[trim={0cm 0cm 0cm 0cm},clip,scale=.25]{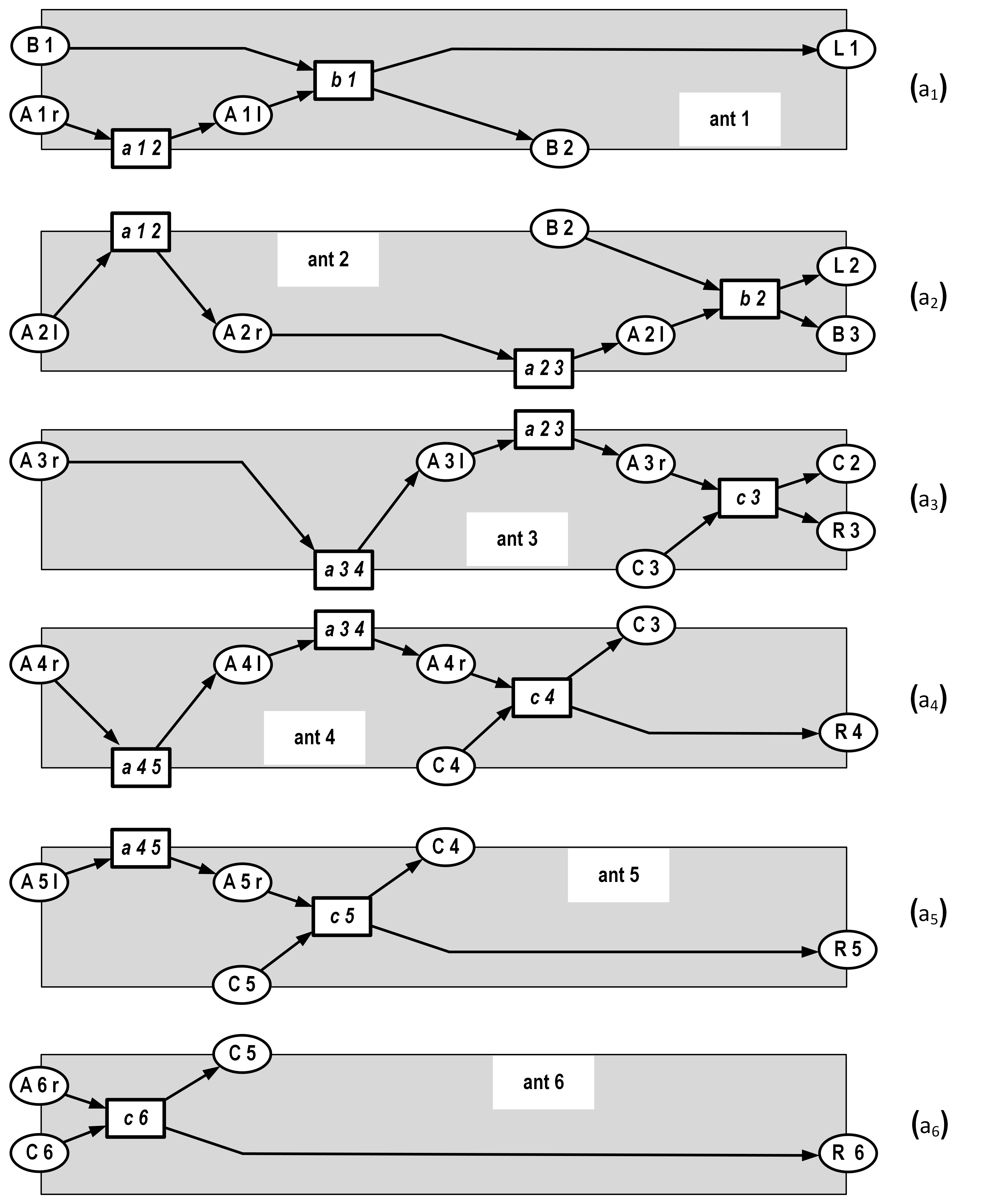}}
\caption{each ant's behavior}
\label{fig:Verhalten_einzeln}
\end{figure}

As each single ant contributes to the run $U$, one may ask for the contribution of each single ant to $U$. Fig.~\ref{fig:Verhalten_einzeln} shows the contribution of ants $1, \dots , 6$, as \textsc{Heraklit} modules, “$\text{ant}\, i$” $(i = 1, \dots, 6)$. For each ant $i$, the module of $i$ starts with zero, one or two events shaped $a\; i\; i\!+\!1$ or $a\; i\!-\!1\; i$, followed by $b$ $i$ or $c$ $i$. This means intuitively, that ant $i$ meets up to two neighbored ants, and then drops down the bar to the left (event $b$ $i$) or to the right (event $c\, i$). Before dropping down, ant $i$ informs its right neighbor ant $i+1$ or its left neighbor ant $i-1$ that it is now the leftmost ant (local state $B$ $i$) or the rightmost ant (local state $C$ $i$).  \textsc{Heraklit} comes with a composition operator “$\bullet$”, such that we can write:
\begin{equation}
U = \text{ant } 1 \bullet \dots \bullet \text{ant } 6.                     \end{equation}

\section{Modeling the bar with ants}
In the definition of runs, replacing temporal order by event order is a fundamental step. It raises the question of how to cope with the new version of runs, how such runs are generated by a kind of system model, etc.  In the rest of this paper, we model the ants system itself, we discuss the composition of system models and runs, and we discuss a schematic representation for ant systems.

\subsection{The model of the ants system}
\begin{figure}[!tb]
\centering
{\includegraphics[trim={0cm 0cm 0cm 0cm},clip,scale=.25]{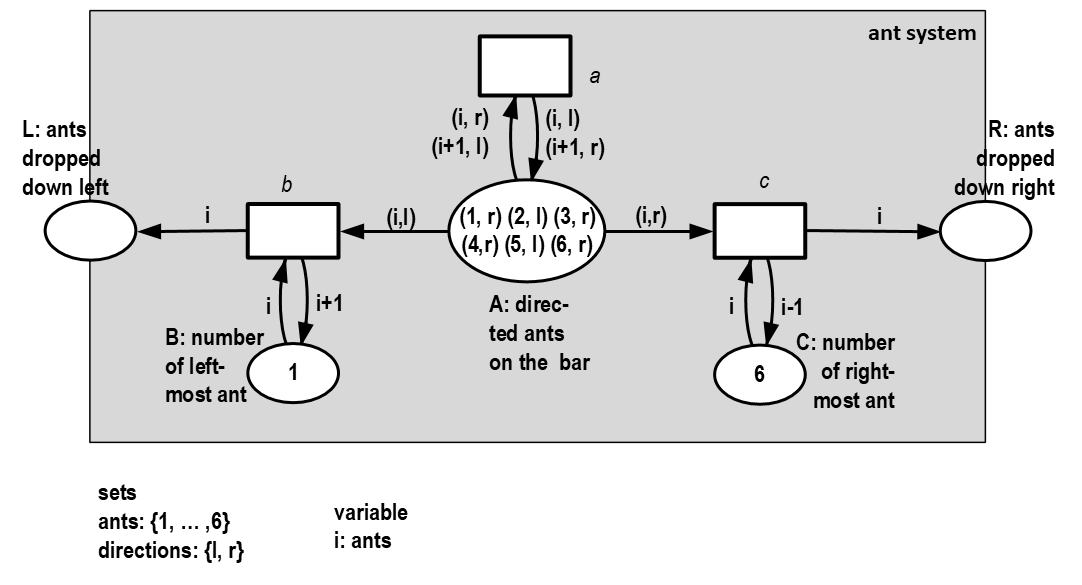}}
\caption{the ants system}
\label{fig:System}
\end{figure}

Fig.~\ref{fig:System} shows the ant system as a module in the \textsc{Heraklit} framework \cite{fettke2021handbook}. The places on its left and right margin are the left and right interface of the module, collecting the dropped down ants. Essentially, this figure shows a high-level Petri net. Its initial marking represents the local state components of the initial state $S_0$. As usual for such Petri nets, in a given marking, a transition is enabled with respect to a \textit{valuation} of the involved variable, $i$. For example, in marking $S_0$, transition “$a$” is enabled with $i = 1$ (representing the event $a\, 1\, 2$), but not with $i = 2$. This way, each transition of Fig.~\ref{fig:System} causes a set of steps. A run (such as in Fig.~\ref{fig:Gesamt-Ablauf}) is a run of the system, if each step of the run can be conceived as a step of the system, and if the initial state of the run fits the initial system state. For Fig.~\ref{fig:System}, this applies to the run $U$ of Fig.~\ref{fig:Gesamt-Ablauf}. Formal details of the notion of a run of a system are presented in \cite{fettke2021handbook}.

\subsection{Composing two bars}
\begin{figure}[!tb]
\centering
{\includegraphics[trim={0cm 0cm 0cm 0cm},clip,scale=.20]{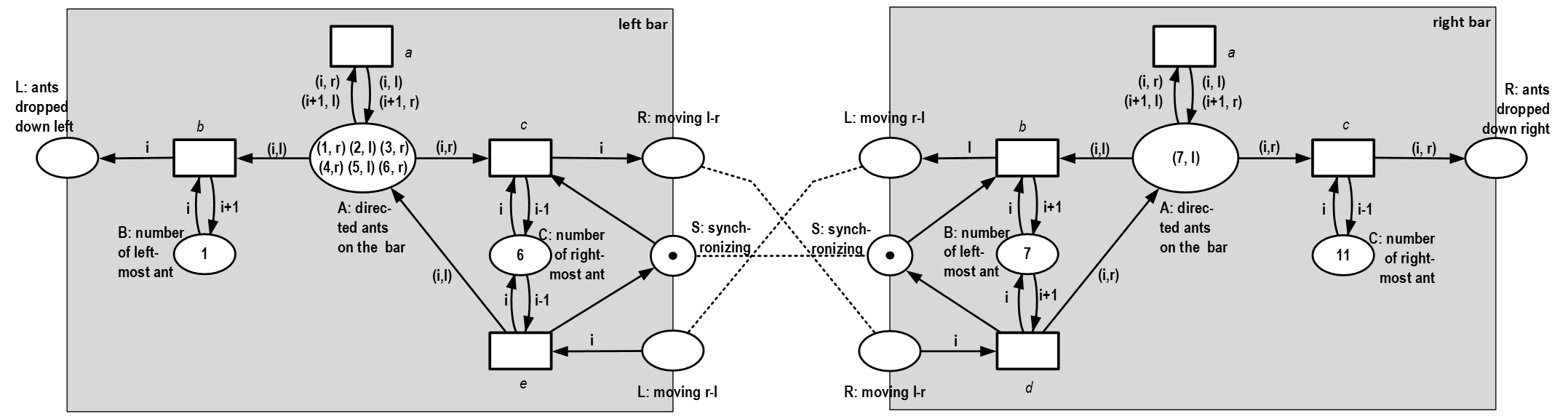}}
\caption{composing two bars}
\label{fig:compose_two}
\end{figure}

As an exercise showing the elegance and technical simplicity of the \textsc{Heraklit} approach, we consider the case of two bars, a left and a right one, linked together. A right moving ant on the left bar no longer drops down to the right of the bar, but moves onto the right bar. Correspondingly, a left moving ant on the right bar no longer drops down to the left of the bar, but moves onto the left bar. 
We model this system with two copies of the above ants system, slightly extending the right interface of the left system, and the left interface of the right system, as shown in Fig.~\ref{fig:compose_two}. Each dotted line links two places that are merged in the composed system. 

The left system of Fig.~\ref{fig:compose_two} extends the ant system of Fig.~\ref{fig:System} by place $L$ and transition $e$. Left-moving ants move from the right to the left bar via the place $L$ and the transition $e$, thus reaching the place $A$ of the left bar. Place $C$ protocols the actually rightmost ant on the bar. In a symmetrical way, the right system of Fig.~\ref{fig:compose_two} introduces the place $R$ and the transition $d$. The synchronizing place $S$ prevents the rightmost ant of the left bar and the leftmost ant of the right bar to slide past each other. In technical terms, $c$ is followed by $d$, before $b$ occurs. Or $b$ is followed by $e$ before $c$ occurs.

Ignoring the dotted lines, the composed system can be written in \textsc{Heraklit} as
\begin{equation}
\text{left bar} \bullet \text{right bar}.
\end{equation}

\begin{figure}[!tb]
\centering
{\includegraphics[trim={0cm 0cm 0cm 0cm},clip,scale=.20]{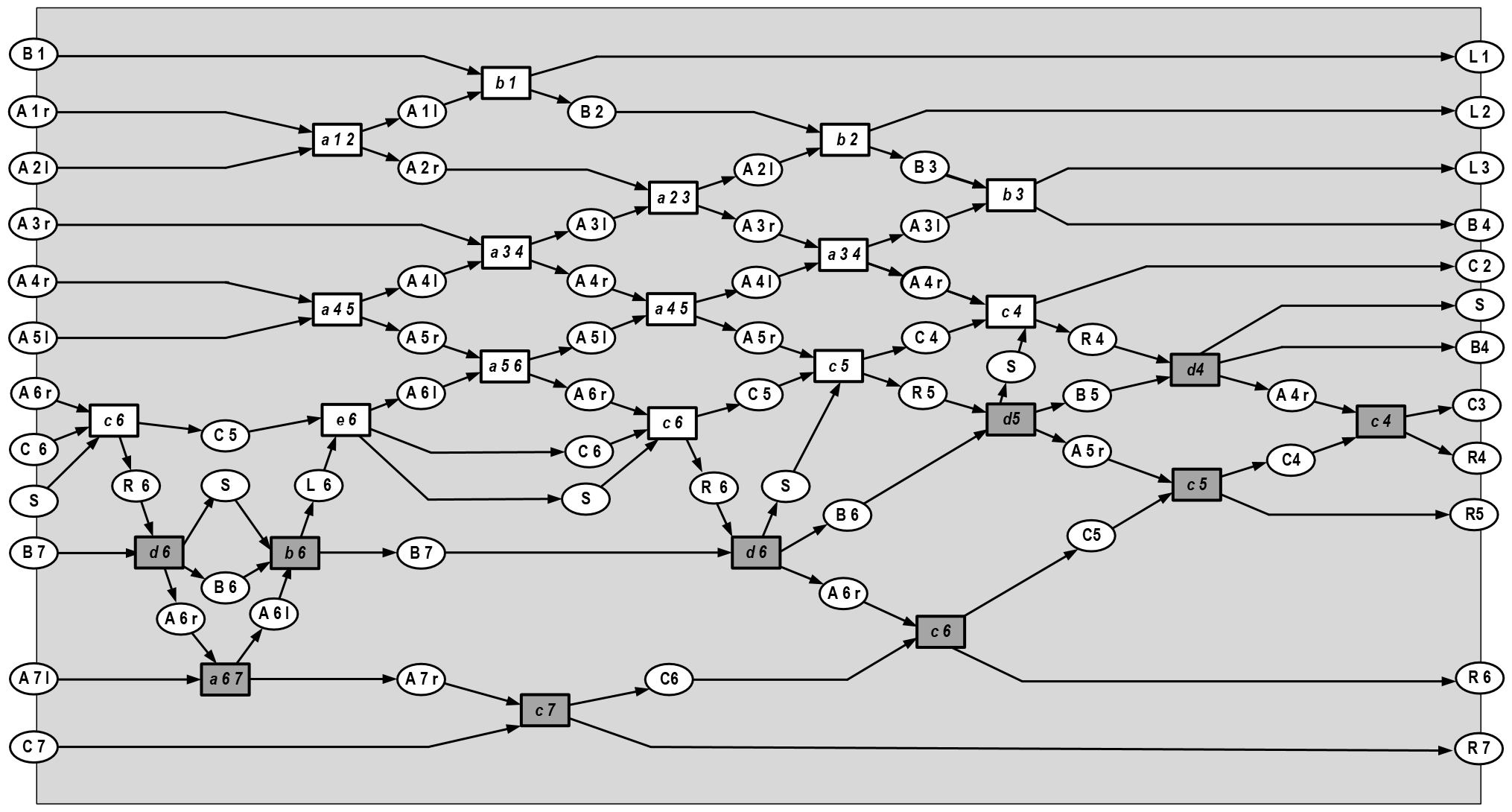}}
\caption{composed run}
\label{fig:composed_run}
\end{figure}

It is interesting to study the composition of runs of the composed system. As initial state, for the left system we assume the state as in Fig.~\ref{fig:Gesamt-Ablauf}. The right system contains initially only one ant, oriented to the left. In the run of Fig.~\ref{fig:composed_run}, ant $6$ moves from the left to the right bar, meeting ant $7$ on the right bar. Both ants swap direction: ant $6$ returns to the left bar, and ant $7$ eventually drops down to the right of the right bar. Ant $6$ meets ant $5$ on the left bar, swaps direction, etc. Eventually, ants $3, \dots, 7$ are dropped down to the right of the right bar. Ants $1$ and $2$ take no notice at all from the newly attached bar, and drop down to the left of the left bar. In Fig.~\ref{fig:composed_run}, the transitions of the right module are shaded.

\subsection{Composing many bars}
\begin{figure}[!tb]
\centering
{\includegraphics[trim={0cm 0cm 0cm 0cm},clip,scale=.25]{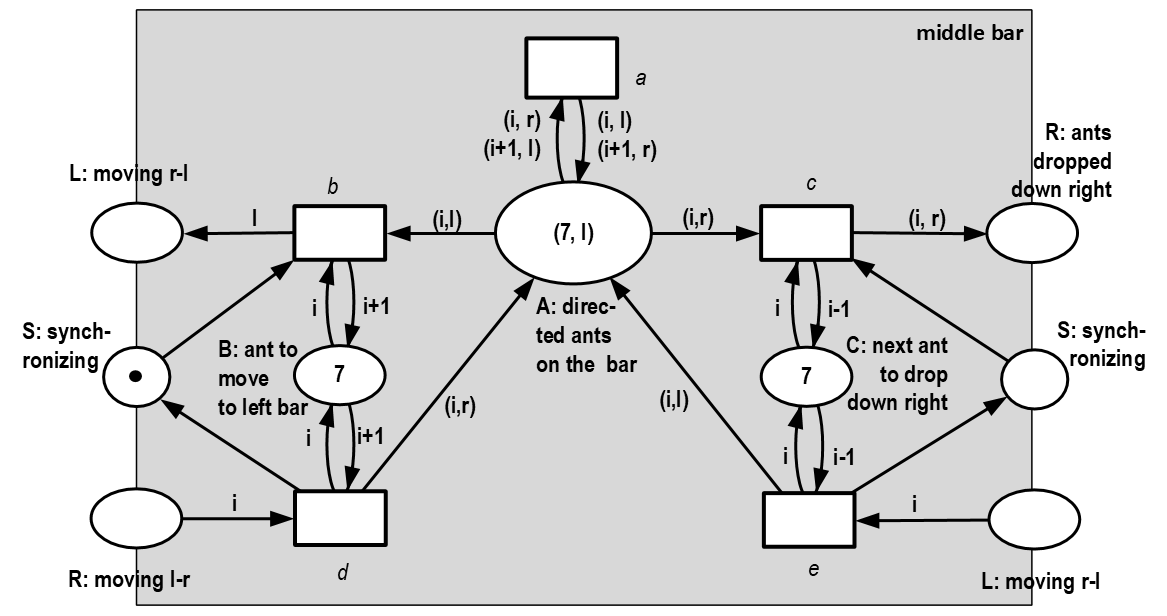}}
\caption{a middle bar}
\label{fig:middle}
\end{figure}

The case of composing two bars can systematically be extended to any number of bars. To this end we employ a middle bar, as in Fig.~\ref{fig:middle}. For example, composition of five bars can be written in \textsc{Heraklit} as 
\begin{equation}
\text{left bar} \bullet \text{middle bar} \bullet \text{middle bar} \bullet \text{middle bar} \bullet \text{right bar}.
\end{equation}

\section{A schematic representation of ants systems}
Fig.~\ref{fig:System} shows an ant system with a fixed initial state, consisting of six ants, each of which is oriented either to the left or to the right. Now we strive at a representation that covers any number of ants, and any orientation of each ant.
We achieve this by help of well-established concepts of general algebra \cite{sanella20212algebraic}, as used in algebraic specifications such as \textit{CASL}, \textit{VDM}, \textit{Z}, etc. The inscriptions of Fig.~\ref{fig:System} represent elements of a set of ants and a set of directions 

\begin{figure}[!tb]
\centering
{\includegraphics[trim={0cm 0cm 0cm 0cm},clip,scale=.25]{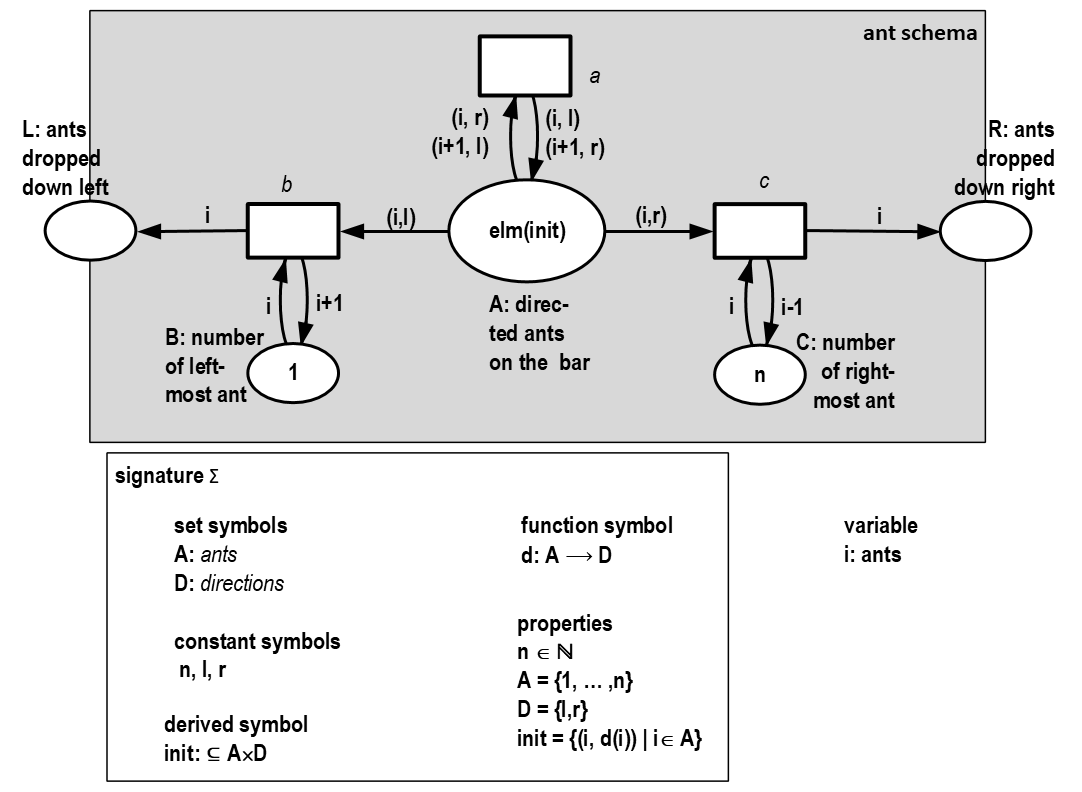}}
\caption{the ant schema}
\label{fig:schema}
\end{figure}

Fig.~\ref{fig:schema} includes a signature, $\Sigma$, some typed symbols for sets, constants and functions, and four properties. As usual in algebraic specifications, an instantiation of $\Sigma$ assigns each symbol a corresponding set, constant, or function, such that the required properties are respected. Thus, an instantiation of $\Sigma$ is structure, denoted as a $\Sigma$–structure. For example, Fig.~\ref{fig:System} includes an instantiation of the signature $\Sigma$ of Fig.~\ref{fig:schema}: The symbols $A$ and $D$ are instantiated by the sets $\{1, \dots , 6\}$ and $\{l, r\}$. The constant symbol $n$ is instantiated by the integer $6$. The symbols $l$ and $r$ are instantiated by themselves. The instantiation of the function symbol $d$ is a function that is implicitly given by the initial marking of the place $A$. Summing up, each instantiations of $\Sigma$ specifies an ant system.

The initial marking of the place $A$ deserves particular attention. One may be tempted to use the symbol “init” as symbolic initial marking. According to the signature $\Sigma$, an instantiation would instantiate “init” as a set ants of directed ants, as one initial token. This is, however, not what we want. Instead, we want each element of the set ants to be a token. In logical terms, with the place $A$ denoting the predicate “directed ants on the bar”, we want not to state ”A(ants)”, but 
\begin{equation}
\forall a \in \text{ants}: A(a).
\end{equation}
Using the \textsc{Heraklit} framework, we denote this by the inscription “elm(init)“ in the place $A$.

\section{Related work}
On the background of classical automata theory and transition systems, a run (i.e. a single behavior) of a discrete system is a set of activities, totally ordered along the evolution of time. Each activity updates a (global) state, with a state containing all what is the case at a distinguished point in time. C.A. Petri challenged this “interleaving semantics” since the 1960ies, and suggested non-sequential processes as a model for single runs of any kind of discrete systems \cite{petri1977non_sequential}. In this view, a run is a set of activities, partially ordered by causal dependencies. Such partial orders have been suggested for many models, in particular for interacting sequential systems. Typical such contributions include \cite{lamport1978time,reisig1984partial,degano1988distributed,koutny1994partial,korff1998true} and may others. But partial order semantics never prevailed. Its technical costs were considered to outperform the gained insight.

The \textsc{Heraklit} framework with its composition operators used above, however reduces the technical burden drastically and reveals further insight. In particular, composition of systems yields exponentially many interleaving runs, but only quadratically many partially ordered runs. The number of runs of a composed system $A \bullet B$ is in general exponentially bigger than the number of runs of $A$ and $B$. The corresponding number of partially ordered runs is only quadratically bigger.

Our discussion of weak orderings in Sec.~\ref{sec:2.6} reflects the discussion of measurement in several scientific fields as discussed by e.g. \cite{krantz1971foundations}: typically it is argued that measurement means that attributes of objects, substances, and events can reasonably be represented numerically such that the observed order of objects etc. is preserved in the numerical representation. In other words, any reasonable kind of measurement yields totally ordered results if different objects has to be represented by different numbers; and weakly ordered results if different objects can be represented by the same number. So, it comes without surprise that the behavioral models of Sec.~\ref{sec:2.1} to \ref{sec:2.4} define weak orderings.

\section{Conclusion}
With this paper, we intend to raise a number of questions, without claiming full-fledged answers: 

\begin{itemize}
\item The fundamental question: Is it possible for each system with timing aspects, to separate time and functionality? If yes, does this separation always yield better insight into the system, or better methods to prove aspects of correctness? If no, can properties of systems be identified, that would characterize this distinction?

\item To which extent does our chosen modeling technique, \textsc{Heraklit}, limit or bias the representation of functionality? 

\item How is the notion of time in real world systems related to the treatment of time in models, and in implementations: In informatics we frequently tend to assume continuous or discrete time to “exist” and to be “measurable” without much of effort. Sections \ref{sec:2.2} to \ref{sec:2.4} show that this assumption is not justified, and the above question is far from trivial.  Before discussing e.g. meeting points of ants with different velocities, it must be clarified how those notions will be fixed in the model. 
\end{itemize}

We suggest to base this kind of questions on the notion of causality. This offers a larger degree of mathematical expressiveness, because causality orders the events of a run not necessarily by a weak ordering (as time based models usually do), but by a more general partial order. We suggest this as a beginning of a theoretical framework for any kind of modeling of collective adaptive systems. This implies new development processes, property preserving verification, refinement, composition etc.

\subsubsection*{Acknowledgment}
We deeply appreciate the careful and thoughtful hints and comments of the referees.

%
%
%

\bibliographystyle{splncs04}
\bibliography{main}





\end{document}